%% file: archead_preprint.tex
\pdfoutput=1
\documentclass[11pt]{article}

\usepackage[preprint]{acl}
\usepackage{times}
\usepackage{latexsym}
\usepackage[T1]{fontenc}
\usepackage[utf8]{inputenc}
\usepackage{microtype}
\usepackage{graphicx}
\usepackage{booktabs}
\usepackage{amsmath}
\usepackage{amssymb}
\usepackage{algorithm}
\usepackage{algorithmic}
\usepackage{placeins}

\title{ARCHead: Activation-Metric Residual Correction\\
for Large Language Model Output Heads}

\author{
Şuayp Talha Kocabay \\
Independent Researcher \\
{\small\texttt{kocabaysuayptalha08@gmail.com}}
\And
Talha Rüzgar Akkuş \\
Independent Researcher \\
{\small\texttt{talharuzgarakkus@gmail.com}}
\AND
Kamer Ali Yüksel \\
aiXplain, Inc. \\
{\small\texttt{kamer@aixplain.com}}
}

\hypersetup{
  pdftitle={ARCHead: Activation-Metric Residual Correction for Large Language Model Output Heads},
  pdfauthor={Şuayp Talha Kocabay, Talha Rüzgar Akkuş, Kamer Ali Yüksel}
}

\begin{document}
\maketitle

\begin{abstract}
Weight-only quantization substantially reduces the storage of large language model (LLM) transformer blocks, but practical backends often retain the final language-modeling head (LM-head) in BF16 or FP16. Quantizing this projection naively can strongly perturb the vocabulary-logit distribution. We present ARCHead, a packed LM-head compressor that combines a quantized low-rank core, group-wise INT4 residuals, and a low-rank correction fitted in an activation-derived metric. ARCHead stores no dense BF16 head and reduces persistent LM-head storage by 3.7--3.9$\times$. On Qwen3-8B-Base, it uses 25.6\% of BF16 head storage while attaining 1.007 relative perplexity; storage-matched naive INT4 yields 1.14--1.16. Replacing the BF16 head left by AWQ or bitsandbytes adds only 0.006--0.007 cross-entropy, with less than 2\% throughput change in our measurements. ARCHead therefore complements block quantizers by compressing the large output projection they can leave untouched. Code is available at \url{https://github.com/suayptalha/archead}.
\end{abstract}

\section{Introduction}

Post-training quantization (PTQ) methods such as GPTQ \citep{gptq}, AWQ \citep{awq}, and bitsandbytes NF4 \citep{qlora} make large language models (LLMs) substantially easier to deploy. Their practical implementations, however, often focus on transformer-block weights and retain the final language-modeling head (LM-head) in BF16 or FP16. This head maps the final hidden state directly to a logit for every vocabulary item. Unlike errors inside a transformer block, an LM-head error is not followed by normalization, a residual path, or another learned transformation; it reaches the softmax directly.

This matters increasingly for large-vocabulary models. The BF16 output projections of Qwen3-8B-Base \citep{qwen} and Gemma-4-E4B \citep{gemma} occupy approximately 1.18 and 1.28\,GB, respectively. Once transformer blocks have been reduced to four bits, such a projection can become one of the largest remaining dense tensors. In our inspection of quantized Qwen3-8B-Base checkpoints, both AWQ and bitsandbytes NF4 retained \texttt{lm\_head} as a BF16 \texttt{Linear} layer of shape $151{,}936\times4{,}096$.

Naive low-bit head quantization is not an adequate remedy. It minimizes a weight-space reconstruction error that treats every hidden direction equally, even though the final hidden states occupy an anisotropic distribution. A small error along a frequently activated direction can alter many vocabulary logits, whereas a larger error along a nearly inactive direction may have little effect. The relevant objective is therefore the expected output error induced by the hidden-state distribution.

We introduce \textbf{ARCHead}, a specialized, packed LM-head compressor. ARCHead first represents the head with quantized low-rank factors and a group-wise INT4 residual. It then approximates the remaining error with a low-rank branch fitted after an activation-derived metric transform. The resulting module is a drop-in output head and stores no dense BF16 copy of the original matrix.

Our contributions are:
\begin{itemize}
    \item We formulate output-head compression in an activation-derived metric and show that, for a fixed quantized core, the ARCHead correction is the best rank-$r$ approximation in that metric before factor quantization.
    \item We develop a packed representation whose measured state-dictionary footprint is 25--27\% of the corresponding BF16 head across five model families.
    \item We show that ARCHead avoids the large quality loss of storage-matched naive INT4, generalizes across three output heads, and can compress the BF16 head left by AWQ and bitsandbytes at a small additional loss.
    \item We evaluate logit fidelity, downstream accuracy, calibration sensitivity, construction time, and generation throughput, and separate persistent parameter storage from backend-dependent runtime peak memory.
\end{itemize}

\section{Related Work}

\paragraph{LLM quantization.}
GPTQ applies second-order information during sequential weight quantization \citep{gptq}; AWQ protects activation-salient channels \citep{awq}; and SmoothQuant moves quantization difficulty between activations and weights \citep{smoothquant}. Other approaches optimize quantization parameters \citep{omniquant}, use layer-wise distillation \citep{zeroquant}, isolate outliers \citep{spqr,squeezellm}, or introduce incoherence transforms and rotations \citep{quip,quipsharp,quarot,spinquant}. AQLM uses additive codes for extreme compression \citep{aqlm}, while LLM.int8() and NF4 provide widely used mixed-precision representations \citep{llmint8,qlora}. These methods primarily target transformer-block linear layers; ARCHead instead addresses the large output projection that practical pipelines may leave dense.

\paragraph{Low-rank and residual correction.}
Low-rank language-model compression \citep{svd_compression} and parameter-efficient low-rank updates \citep{lora} demonstrate that structured matrix components can be represented compactly. ARCHead differs in both target and objective: it computes a low-rank approximation of the \emph{quantization residual} after transforming that residual by a metric estimated from LM-head inputs. This directly allocates correction capacity to directions that affect the observed logits.

\paragraph{Realized storage.}
An intended bit width does not itself establish deployable compression: an implementation can retain a dequantized tensor or materialize one in its serialized state. We therefore measure bytes from the actual tensors registered by the packed head and report persistent storage separately from temporary forward-pass memory.

\section{Method}

\subsection{Activation-Metric Objective}

Let the dense output-head weight be $\mathbf{W}\in\mathbb{R}^{V\times D}$ and let $\mathbf{H}\in\mathbb{R}^{N\times D}$ contain final hidden states from $N$ calibration tokens. For an approximation $\widehat{\mathbf{W}}$, write $\boldsymbol{\Delta}=\mathbf{W}-\widehat{\mathbf{W}}$. The empirical squared logit error is
\begin{align}
\mathcal{E}
&=\frac{1}{N}\lVert\mathbf{H}\boldsymbol{\Delta}^{\top}\rVert_F^2
=\operatorname{Tr}\!\left(\boldsymbol{\Delta}\mathbf{C}
\boldsymbol{\Delta}^{\top}\right),
\label{eq:logit_error}
\end{align}
where $\mathbf{C}=\mathbf{H}^\top\mathbf{H}/N$. A Frobenius weight objective is the special case $\mathbf{C}=\mathbf{I}$ and ignores the activation geometry.

ARCHead uses a damped metric transform. With
$\mathbf{C}_\lambda=\mathbf{C}+\lambda\bar{c}\mathbf{I}
=\mathbf{Q}\boldsymbol{\Lambda}\mathbf{Q}^\top$,
where $\bar{c}$ is the mean diagonal of $\mathbf{C}$, define
\begin{equation}
\mathbf{T}_p=\mathbf{Q}\boldsymbol{\Lambda}^{p}\mathbf{Q}^\top,
\qquad
\mathbf{T}_p^{-1}=\mathbf{Q}\boldsymbol{\Lambda}^{-p}\mathbf{Q}^\top .
\end{equation}
At $p=\tfrac12$, the induced objective is the damped form of Eq.~\ref{eq:logit_error}; other values adjust how strongly dominant activation directions are emphasized. We select $p$ on calibration data and use $p=0.75$ for Qwen3-8B-Base.

\subsection{Core and Residual Correction}

\begin{figure}[t]
\centering
\includegraphics[width=\columnwidth]{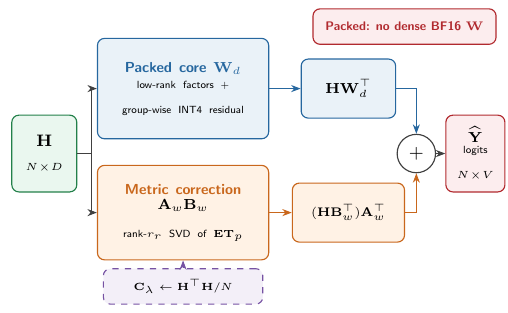}
\caption{ARCHead decomposes the dense output head into a packed quantized core $\mathbf{W}_d$ and an activation-metric low-rank correction. The original BF16 matrix is discarded after construction.}
\label{fig:method}
\end{figure}

The quantized core is
\begin{equation}
\mathbf{W}_d =
Q_5(\mathbf{A}_c)Q_8(\mathbf{B}_c)
+Q_{\mathrm{SC4}}\!\left(\mathbf{W}-\mathbf{A}_c\mathbf{B}_c\right),
\end{equation}
where $\mathbf{A}_c\mathbf{B}_c$ is a rank-$r_c$ approximation, $Q_b$ denotes $b$-bit group quantization, and SC4 denotes the packed signed-INT4 residual with quantized group scales. The residual left by this core is $\mathbf{E}=\mathbf{W}-\mathbf{W}_d$.

ARCHead computes a rank-$r_r$ randomized truncated SVD
\begin{equation}
\mathbf{E}\mathbf{T}_p \approx
\mathbf{U}_{r_r}\boldsymbol{\Sigma}_{r_r}\mathbf{V}_{r_r}^{\top}
\end{equation}
and maps the right factor back:
\begin{equation}
\mathbf{A}_w=\mathbf{U}_{r_r}\boldsymbol{\Sigma}_{r_r},
\qquad
\mathbf{B}_w=\mathbf{V}_{r_r}^{\top}\mathbf{T}_p^{-1}.
\end{equation}
The final approximation and logits are
\begin{align}
\widehat{\mathbf{W}} &= \mathbf{W}_d+\mathbf{A}_w\mathbf{B}_w,\\
\widehat{\mathbf{Y}} &= \mathbf{H}\mathbf{W}_d^\top+
(\mathbf{H}\mathbf{B}_w^\top)\mathbf{A}_w^\top .
\end{align}
For the packed ARCHead variant, both correction factors are stored with row/group-wise INT8 quantization.

\paragraph{Conditional optimality.}
For fixed $\mathbf{W}_d$ and invertible $\mathbf{T}_p$, the unquantized factors above minimize
\begin{equation}
\left\lVert(\mathbf{E}-\mathbf{A}\mathbf{B})\mathbf{T}_p\right\rVert_F^2
\end{equation}
over all rank-$r_r$ products $\mathbf{A}\mathbf{B}$. This follows directly from the Eckart--Young--Mirsky theorem after changing variables to $\widetilde{\mathbf{E}}=\mathbf{E}\mathbf{T}_p$. The guarantee is conditional on the fixed core and applies before correction-factor quantization.

\paragraph{Proof.}
Set $\widetilde{\mathbf{B}}=\mathbf{B}\mathbf{T}_p$. Because $\mathbf{T}_p$ is invertible,
$\operatorname{rank}(\mathbf{A}\widetilde{\mathbf{B}})=\operatorname{rank}(\mathbf{A}\mathbf{B})$, so the problem is equivalent to
\begin{equation}
\min_{\operatorname{rank}(\mathbf{A}\widetilde{\mathbf{B}})\le r_r}
\left\lVert\widetilde{\mathbf{E}}-\mathbf{A}\widetilde{\mathbf{B}}\right\rVert_F^2.
\end{equation}
By the Eckart--Young--Mirsky theorem, the minimizer is the rank-$r_r$ truncated SVD $\mathbf{U}_{r_r}\boldsymbol{\Sigma}_{r_r}\mathbf{V}_{r_r}^\top$ of $\widetilde{\mathbf{E}}$. Choosing $\mathbf{A}=\mathbf{U}_{r_r}\boldsymbol{\Sigma}_{r_r}$ and $\mathbf{B}=\mathbf{V}_{r_r}^\top\mathbf{T}_p^{-1}$ yields the stated factors. For $p=\tfrac12$, the objective is the damped empirical logit MSE. Our randomized truncated SVD and subsequent factor quantization approximate this ideal correction, so the result does not assert global optimality of the quantized end-to-end module.

\subsection{Packed Representation and Integration}

The packed \texttt{ARCHead} registers only the SC4 residual and scales, quantized core factors, quantized correction factors, and shape metadata. Its persistent size is measured directly as
\begin{equation}
\begin{aligned}
\operatorname{bytes}(\mathrm{ARCHead})
&=\sum_i \operatorname{numel}(\theta_i) \\
&\quad{}\times \operatorname{element\_size}(\theta_i),
\end{aligned}
\end{equation}
over all parameters and buffers $\theta_i$ in the serialized module. The original $V\times D$ BF16 matrix is not registered. During model conversion, the dense head is used to construct ARCHead and is then discarded. The module can replace the output embedding after block quantization, so ARCHead is complementary to AWQ, bitsandbytes, or another block backend. Algorithm~\ref{alg:compression} gives the complete construction procedure.

\paragraph{State-dictionary accounting.}
The packed implementation is not merely a theoretical bit-counting scheme: unlike prototypes that retain dequantized dense tensors at load time, it stores no dense BF16 $V\times D$ head. Dense storage is $VD\times2$ bytes, whereas packed storage is the measured sum above over the registered \texttt{state\_dict}. The stored components are the packed INT4 residual and its FP16/BF16 scales; the 5-bit left and 8-bit right core factors and their scales; row/group-wise INT8 correction factors $\mathbf{A}_w$ and $\mathbf{B}_w$; quantization scales and zero-point metadata; and group-size and shape metadata. Table~\ref{tab:packed_storage_detailed} reports values measured from instantiated PyTorch buffers rather than theoretical bit widths.

\begin{table*}[t]
\centering
\begin{tabular}{lrrrrr}
\toprule
Model & Vocab & Hidden & Dense BF16 MB & Packed ARCHead MB & Compression \\
\midrule
Qwen3-8B-Base & 151,936 & 4,096 & 1187.0 & 303.96 & 3.905$\times$ \\
Gemma-4-E4B & 262,144 & 2,560 & 1280.0 & 345.41 & 3.706$\times$ \\
VibeThinker-3B & 151,936 & 2,048 & 593.5 & 153.23 & 3.873$\times$ \\
Mistral-7B-v0.3 & 32,768 & 4,096 & 256.0 & 65.61 & 3.902$\times$ \\
LFM2.5-8B-A1B & 128,000 & 2,048 & 500.0 & 129.10 & 3.873$\times$ \\
\bottomrule
\end{tabular}
\caption{Detailed packed \texttt{ARCHead} storage validation, calculated from the instantiated PyTorch \texttt{state\_dict} buffers.}
\label{tab:packed_storage_detailed}
\end{table*}

\paragraph{Load-time versus peak memory.}
Persistent LM-head memory measures the parameter allocation when loading the model onto the GPU and is the appropriate quantity for evaluating the stored output-head footprint. Table~\ref{tab:persistent_memory} reports the corresponding savings. Forward-pass peak memory is separate: it includes activations, output logits, temporary buffers, CUDA allocator behavior, backend workspaces, and the KV cache. ARCHead reduces persistent output-head parameters; we do not conflate this with backend- and workload-dependent runtime peak memory.

\begin{table}[t]
\centering
\resizebox{\columnwidth}{!}{
\begin{tabular}{lrrrr}
\toprule
Model & Dense MB & ARCHead MB & Saved MB & Ratio \\
\midrule
Qwen3-8B & 1188.0 & 304.0 & 884.0 & 3.91$\times$ \\
Gemma-4-E4B & 1280.0 & 345.4 & 934.6 & 3.71$\times$ \\
VibeThinker-3B & 594.0 & 153.4 & 440.6 & 3.87$\times$ \\
Mistral-7B-v0.3 & 256.0 & 65.6 & 190.4 & 3.90$\times$ \\
LFM2.5-8B & 500.0 & 130.1 & 369.9 & 3.84$\times$ \\
\bottomrule
\end{tabular}}
\caption{Load-time persistent LM-head memory savings.}
\label{tab:persistent_memory}
\end{table}

\paragraph{Peak-memory measurement protocol.}
Before measurement, we execute \texttt{torch.cuda.empty\_cache()}, reset PyTorch peak-memory statistics, and synchronize the CUDA device before and after execution. We measure and report load-time persistent parameter memory separately from the forward-pass peak, which inherently includes dynamic logits, temporary buffers, allocator state, and backend workspaces; consequently, we avoid unsupported claims about backend-dependent peak-memory reductions.

\subsection{Construction Algorithm}
\label{sec:algorithm}

The inputs are the dense LM-head weight $\mathbf{W}\in\mathbb{R}^{V\times D}$, hidden calibration activations $\mathbf{H}\in\mathbb{R}^{N\times D}$, core and correction ranks $r_c$ and $r_r$, group size $g$, metric power $p$, and damping $\lambda$. The procedure first estimates the damped activation covariance $\mathbf{C}=\mathbf{H}^\top\mathbf{H}/N+\lambda\mathbf{I}$. It then constructs the quantized core $\mathbf{W}_d$ by quantizing a low-rank approximation of $\mathbf{W}$ and applying group-wise low-bit quantization to the remaining residual.

Next, ARCHead computes the core error $\mathbf{E}=\mathbf{W}-\mathbf{W}_d$, eigendecomposes $\mathbf{C}=\mathbf{Q}\boldsymbol{\Lambda}\mathbf{Q}^\top$, and forms $\mathbf{T}_p=\mathbf{Q}\boldsymbol{\Lambda}^p\mathbf{Q}^\top$ and its inverse. The transformed residual $\widetilde{\mathbf{E}}=\mathbf{E}\mathbf{T}_p$ is truncated to rank $r_r$; mapping its right factor through $\mathbf{T}_p^{-1}$ produces $\mathbf{A}_w$ and $\mathbf{B}_w$. We store $\mathbf{A}_w$ with row-wise INT8 quantization and $\mathbf{B}_w$ with group-wise INT8 quantization, pack every component, and discard the dense BF16 tensor. Inference then evaluates $\mathbf{Y}=\mathbf{H}\mathbf{W}_d^\top+(\mathbf{H}\mathbf{B}_w^\top)\mathbf{A}_w^\top$.

\begin{algorithm}[t]
\caption{ARCHead compression}
\label{alg:compression}
\begin{algorithmic}[1]
\renewcommand{\algorithmicrequire}{\textbf{Input:}}
\renewcommand{\algorithmicensure}{\textbf{Output:}}
\REQUIRE Dense head $\mathbf{W}$, activations $\mathbf{H}$, ranks $r_c,r_r$, group size $g$, metric power $p$, damping $\lambda$
\ENSURE Packed \texttt{ARCHead} module
\STATE $\mathbf{C}\leftarrow\mathbf{H}^\top\mathbf{H}/N+\lambda\mathbf{I}$
\STATE $\mathbf{W}_d\leftarrow\text{QuantizeCore}(\mathbf{W},r_c,g)$
\STATE $\mathbf{E}\leftarrow\mathbf{W}-\mathbf{W}_d$
\STATE $\mathbf{Q},\boldsymbol{\Lambda}\leftarrow\text{Eigendecompose}(\mathbf{C})$
\STATE $\mathbf{T}_p\leftarrow\mathbf{Q}\boldsymbol{\Lambda}^p\mathbf{Q}^\top$
\STATE $\widetilde{\mathbf{E}}\leftarrow\mathbf{E}\mathbf{T}_p$
\STATE $\mathbf{U},\boldsymbol{\Sigma},\mathbf{V}\leftarrow\text{TruncatedSVD}(\widetilde{\mathbf{E}},r_r)$
\STATE $\mathbf{A}_w\leftarrow\mathbf{U}\boldsymbol{\Sigma}$; $\mathbf{B}_w\leftarrow\mathbf{V}^\top\mathbf{T}_p^{-1}$
\STATE $(\mathbf{A}_w,\mathbf{B}_w)\leftarrow\text{QuantizeINT8}(\mathbf{A}_w,\mathbf{B}_w)$
\RETURN $\text{PackBuffers}(\mathbf{W}_d,\mathbf{A}_w,\mathbf{B}_w)$
\end{algorithmic}
\end{algorithm}

\section{Experimental Setup}

\paragraph{Models.}
We evaluate head quality on Qwen3-8B-Base \citep{qwen}, Gemma-4-E4B \citep{gemma}, and VibeThinker-3B \citep{vibethinker}. Their $(V,D)$ dimensions are $(151{,}936,4096)$, $(262{,}144,2560)$, and $(151{,}936,2048)$. Packed-storage validation additionally includes Mistral-7B-v0.3 \citep{mistral} and LFM2.5-8B-A1B \citep{lfm}.

\paragraph{Data and metrics.}
We collect calibration activations from the training split of WikiText-103 \citep{wikitext} and evaluate cross-entropy (CE) on 16,384 held-out test tokens. We report perplexity (PPL), relative PPL (compressed PPL divided by dense PPL), $\Delta$CE, logit MSE, KL divergence, dense-prediction agreement, and persistent head bytes. For the GPTQ-head comparison, evaluation activations are fixed and disjoint from calibration; each 2K--16K budget uses three matched calibration seeds.

\paragraph{Baselines and implementation.}
Head-only baselines are BF16, row-wise INT8, group INT4, SVD8+INT4, and a GPTQ-style INT4 output head. Transformer blocks remain BF16 unless a row explicitly names AWQ or bitsandbytes NF4. On Qwen, ARCHead uses $r_c=10$, $r_r=6$, groups of 64, $p=0.75$, and ridge $10^{-3}$. Primary deployment measurements use one NVIDIA RTX Pro 6000.

\paragraph{Reproducibility protocol.}
We use the checkpoints \texttt{Qwen/Qwen3-8B-Base}, \texttt{google/gemma-4-E4B}, \texttt{WeiboAI/VibeThinker-3B}, \texttt{mistralai/Mistral-7B-v0.3}, and \texttt{LiquidAI/LFM2.5-8B-A1B}. Calibration activations come from the training split of \texttt{Salesforce/wikitext-103-raw-v1}; quality is evaluated on a fixed 16,384-token test set. The matched GPTQ-head comparison uses 2,048, 4,096, 8,192, and 16,384 calibration tokens with seeds 0, 1, and 2; the broader ARCHead sensitivity sweep uses 4,096--65,536 tokens. AutoAWQ and bitsandbytes are used for the full-model hybrid experiments. GPTQ is evaluated as a standalone head-only INT4 baseline with the same activation cache and evaluation protocol as ARCHead; it is omitted from the hybrid experiments because of backend compatibility constraints. Construction times are measured on the same NVIDIA RTX Pro 6000 and reported as paired relative speedups. Code is available at \url{https://github.com/suayptalha/archead}.

\section{Results}

\subsection{Head-Only Quality}

\begin{table}[t]
\centering
\small
\begin{tabular}{lrrrr}
\toprule
Method & Ratio & CE & $\Delta$CE & Rel.\ PPL \\
\midrule
Dense BF16 & 1.000 & 2.443 & 0.000 & 1.000 \\
Row INT8 & 0.500 & 2.460 & +0.017 & 1.017 \\
Group INT4 & 0.266 & 2.583 & +0.140 & 1.151 \\
SVD8+INT4 & 0.270 & 2.634 & +0.191 & 1.211 \\
\textbf{ARCHead} & \textbf{0.256} & \textbf{2.450} & \textbf{+0.007} & \textbf{1.007} \\
\bottomrule
\end{tabular}
\caption{Qwen3-8B-Base head-only results on WikiText-103. Ratio is persistent head storage relative to BF16.}
\label{tab:qwen_head_only}
\end{table}

At nearly the same storage, Group INT4 raises relative PPL to 1.151, and adding an ordinary low-rank component does not repair the loss (Table~\ref{tab:qwen_head_only}). ARCHead instead remains near the dense head at 1.007 relative PPL. The ablation in Section~\ref{sec:ablation} shows that the improvement does not come from the quantized core alone.

Across architectures, ARCHead obtains relative PPL values of 1.007, 1.010, and 1.027 on Qwen, Gemma, and VibeThinker, respectively (Table~\ref{tab:cross_model}). The gain over naive INT4 is largest on Qwen and narrower on Gemma, whose group-INT4 head is already strong. This variation argues against assuming that every output head has the same sensitivity.

\begin{table}[t]
\centering
\footnotesize
\begingroup
\setlength{\tabcolsep}{4pt}
\begin{tabular}{@{}lrrr@{}}
\toprule
Model & INT4 R-PPL & ARCHead R-PPL & Ratio \\
\midrule
Qwen3-8B & 1.151 & \textbf{1.007} & 0.256 \\
Gemma-4-E4B & 1.013 & \textbf{1.010} & 0.269 \\
VibeThinker-3B & 1.050 & \textbf{1.027} & 0.257 \\
\bottomrule
\end{tabular}
\endgroup
\caption{Head-only results across model families. The bold entry marks the lower relative PPL when the displayed rounded values differ materially.}
\label{tab:cross_model}
\end{table}

\subsection{Compressing Heads Left by Block Quantizers}

AWQ and bitsandbytes NF4 both retained a 1.18\,GB BF16 head in our Qwen checkpoint. Replacing only this head with ARCHead reduced its stored size to 25.6\% of BF16. The extra CE over each already-quantized model was 0.006 for AWQ and 0.007 for NF4 (Table~\ref{tab:hybrid}). ARCHead is not claimed to improve the block quantizer; it recovers most of the persistent storage occupied by the dense output projection at a small additional loss.

\paragraph{Backend inspection.}
We verified the retained head directly. Under bitsandbytes NF4, \texttt{lm\_head} remains a \texttt{torch.nn.Linear} with BF16 weights of shape \texttt{[151936, 4096]} on \texttt{cuda:0}. Under AWQ it likewise remains a BF16 \texttt{torch.nn.Linear} of the same shape; it is on CPU immediately after loading and can be moved to CUDA for evaluation. Thus, deployments described as fully quantized can still contain a dense BF16 output projection, and ARCHead operates orthogonally to their transformer-block quantization.

\begin{table}[t]
\centering
\small
\begin{tabular}{llrr}
\toprule
Blocks & Head & Head ratio & Extra CE \\
\midrule
AWQ 4-bit & BF16 & 1.000 & 0.000 \\
AWQ 4-bit & \textbf{ARCHead} & \textbf{0.256} & +0.006 \\
BNB NF4 & BF16 & 1.000 & 0.000 \\
BNB NF4 & \textbf{ARCHead} & \textbf{0.256} & +0.007 \\
\bottomrule
\end{tabular}
\caption{ARCHead replaces the dense head left by two block-quantization backends on Qwen3-8B-Base.}
\label{tab:hybrid}
\end{table}

\subsection{Packed Storage Is Realized}

Summing the tensors actually registered in \texttt{ARCHead} yields 3.71--3.91$\times$ compression across five heads (Table~\ref{tab:packed}). This is a persistent state-dictionary and load-time parameter reduction, not a claim about total forward-pass peak VRAM. The latter also includes logits, activations, allocator state, backend workspaces, and the KV cache.

\begin{table}[t]
\centering
\small
\begingroup
\setlength{\tabcolsep}{3pt}
\begin{tabular}{@{}lrrr@{}}
\toprule
Model & BF16 MB & ARCHead MB & Compression \\
\midrule
Qwen3-8B & 1188.0 & 304.0 & 3.91$\times$ \\
Gemma-4-E4B & 1280.0 & 345.4 & 3.71$\times$ \\
VibeThinker-3B & 594.0 & 153.4 & 3.87$\times$ \\
Mistral-7B-v0.3 & 256.0 & 65.6 & 3.90$\times$ \\
LFM2.5-8B & 500.0 & 130.1 & 3.84$\times$ \\
\bottomrule
\end{tabular}
\endgroup
\caption{Measured persistent LM-head storage. Values are sums over actual packed buffers.}
\label{tab:packed}
\end{table}

\subsection{Comparison with GPTQ-Style Head Quantization}
\label{sec:gptq}

We compare ARCHead with a GPTQ-style INT4 quantizer applied only to the Qwen LM-head. Both methods use the same calibration activations and fixed test activations, and their effective sizes are comparable (25.61\% for ARCHead and 25.78\% for GPTQ). Table~\ref{tab:gptq} reports means and standard deviations over three matched calibration seeds.

\begin{table}[t]
\centering
\resizebox{\columnwidth}{!}{
\begin{tabular}{rrrrrr}
\toprule
& \multicolumn{2}{c}{Relative PPL $\downarrow$} &
\multicolumn{2}{c}{Logit MSE $\downarrow$} & Build \\
\cmidrule(lr){2-3}\cmidrule(lr){4-5}
Tokens & GPTQ & ARCHead & GPTQ & ARCHead & speedup \\
\midrule
2K  & $1.0104{\pm}.0028$ & $\mathbf{1.0058{\pm}.0012}$ & 8,811 & \textbf{7,025} & 2.58$\times$ \\
4K  & $1.0128{\pm}.0018$ & $\mathbf{1.0065{\pm}.0007}$ & 8,245 & \textbf{6,985} & 2.41$\times$ \\
8K  & $1.0093{\pm}.0032$ & $\mathbf{1.0062{\pm}.0004}$ & 7,715 & \textbf{6,964} & 2.40$\times$ \\
16K & $1.0105{\pm}.0007$ & $\mathbf{1.0062{\pm}.0004}$ & 7,349 & \textbf{6,956} & 2.36$\times$ \\
\bottomrule
\end{tabular}}
\caption{Three-seed, head-only comparison on Qwen3-8B-Base. Build speedup is GPTQ time divided by ARCHead time.}
\label{tab:gptq}
\end{table}

ARCHead has lower mean relative PPL and logit MSE at every tested budget. Measured as the excess above dense relative PPL, it reduces GPTQ's degradation by 44.0\%, 49.3\%, 33.6\%, and 40.6\% from 2K through 16K tokens. It also constructs the head 2.36--2.58$\times$ faster in this implementation. At 8K, GPTQ is better on one of three individual seeds, but ARCHead retains the lower mean and markedly lower variance; at the other budgets ARCHead is better on all three seeds.

The methods use activation information differently. GPTQ commits sequential discrete weight updates and propagates their error through an inverse-Hessian factor. ARCHead first fixes a quantized core, then globally allocates rank-$r_r$ correction capacity to the dominant modes of the entire vocabulary-wide residual. The conditional optimum in Section~3.2 explains the specific theoretical advantage of this stage: among rank-$r_r$ corrections to that core, no other unquantized correction has lower error in ARCHead's chosen activation metric. The empirical result shows that this global residual repair is more effective than our storage-matched GPTQ-head implementation in the tested regime; it is not a claim that ARCHead dominates GPTQ for arbitrary layers or settings.

\subsection{Logit Fidelity}

\begin{table}[t]
\centering
\small
\begin{tabular}{lrrrr}
\toprule
Method & Ratio & Top-1 & KL $\downarrow$ & MSE $\downarrow$ \\
\midrule
Row INT8 & 0.500 & 97.56 & 0.0013 & 740 \\
Group INT4 & 0.266 & 76.95 & 0.1671 & 107,599 \\
SVD8+INT4 & 0.267 & 76.11 & 0.1611 & 103,015 \\
\textbf{ARCHead} & \textbf{0.256} & \textbf{93.05} & \textbf{0.0113} & \textbf{7,167} \\
\bottomrule
\end{tabular}
\caption{Fidelity to dense Qwen logits. Top-1 is agreement in percent. ARCHead is compared with storage-matched low-bit baselines; INT8 uses roughly twice its storage.}
\label{tab:logit}
\end{table}

ARCHead preserves the dense top-1 token on 93.05\% of positions, compared with approximately 76\% for the storage-matched baselines (Table~\ref{tab:logit}). It also reduces their KL divergence by more than an order of magnitude. Top-5 agreement reaches 99.95\% and top-10 agreement 100.00\%. Row INT8 remains more accurate, but uses approximately twice the head storage.

\subsection{Ablation: The Correction Is Decisive}
\label{sec:ablation}

\begin{table}[t]
\centering
\small
\begin{tabular}{lrrr}
\toprule
Variant & Ratio & $\Delta$CE & Rel.\ PPL \\
\midrule
Group INT4 & 0.266 & +0.140 & 1.151 \\
SVD8+INT4 & 0.270 & +0.191 & 1.211 \\
ARCHead core only & 0.230 & +0.126 & 1.134 \\
\textbf{ARCHead full} & \textbf{0.256} & \textbf{+0.007} & \textbf{1.007} \\
\bottomrule
\end{tabular}
\caption{Qwen head-only ablation. Removing the activation-metric correction accounts for most of the quality loss.}
\label{tab:ablation}
\end{table}

The quantized core alone yields 1.134 relative PPL (Table~\ref{tab:ablation}). Adding the rank-6 activation-metric correction lowers this to 1.007 for a 2.6-point increase in storage ratio. Thus, the low-rank core is not sufficient: the decisive component is the correction fitted to the core's remaining, activation-weighted error.

\subsection{Downstream Sanity Check}

We evaluate the Qwen3-8B-Base head variants with \texttt{lm-evaluation-harness} \citep{evalharness} on HellaSwag \citep{zellers2019hellaswag}, TruthfulQA MC2 \citep{lin2022truthfulqa}, and WinoGrande \citep{sakaguchi2021winogrande}. Transformer blocks and evaluation settings are fixed across heads. The differences are small, and the suite is too limited to support a claim of downstream improvement; we use it only to check for an obvious regression after replacing the output head.

\begin{table}[t]
\centering
\resizebox{\columnwidth}{!}{
\begin{tabular}{lccc}
\toprule
Method & HellaSwag & TruthfulQA MC2 & WinoGrande \\
\midrule
Dense BF16 & 0.504 & 0.5353 & 0.632 \\
Group INT4 & 0.500 & 0.5310 & 0.648 \\
GPTQ-head INT4 & 0.500 & 0.5364 & 0.640 \\
ARCHead & 0.504 & 0.5408 & 0.644 \\
\bottomrule
\end{tabular}}
\caption{Downstream accuracy sanity check. Small differences should not be interpreted as statistically established task-level gains.}
\label{tab:downstream}
\end{table}

ARCHead matches the dense head on HellaSwag and remains within the small spread among head variants on TruthfulQA MC2 and WinoGrande (Table~\ref{tab:downstream}). These results reveal no obvious regression, but they do not establish task-level gains or preservation of every model capability.

\subsection{Efficiency and Calibration}

Generation throughput changes by less than 2\% in all measured configurations. Qwen throughput is 4468 versus 4467 tokens/s for BF16 and ARCHeads, 3452 versus 3451 with AWQ blocks, and 1789 versus 1813 with NF4 blocks; VibeThinker measures 7091 versus 6947. Small positive differences are treated as measurement noise, not speedups. A preliminary fused-kernel evaluation gives the same conclusion (Appendix~\ref{sec:app_future_kernel}).

ARCHead's Qwen relative PPL remains between 1.0070 and 1.0076 as calibration size varies from 4K to 64K tokens (Table~\ref{tab:calibration}). This suggests that the dominant hidden-space directions can be estimated with a modest sample for this model, but we do not claim calibration-size invariance in general. A few thousand tokens suffice to estimate the covariance accurately in this experiment; this is not a claim of universal calibration independence.

\begin{table}[t]
\centering
\begin{tabular}{rr}
\toprule
Calibration tokens & Relative PPL \\
\midrule
4,096 & $\sim$1.00725 \\
8,192 & $\sim$1.00703 \\
16,384 & $\sim$1.00714 \\
32,768 & $\sim$1.00761 \\
65,536 & $\sim$1.00749 \\
\bottomrule
\end{tabular}
\caption{Calibration sensitivity sweep on Qwen3-8B-Base.}
\label{tab:calibration}
\end{table}

\section{Conclusion}

ARCHead addresses a concrete gap in quantized LLM deployment: the large BF16 output projection that can remain after transformer-block quantization. Its activation-metric residual branch repairs the structured error left by a compact quantized core, while its packed module realizes a measured 3.7--3.9$\times$ persistent head-storage reduction. Across the tested heads, ARCHead provides a practical quality--storage trade-off and can be composed with AWQ or bitsandbytes without materially changing generation throughput. The method is intentionally complementary to full-model quantizers rather than a replacement for them.

\section*{Limitations}

ARCHead is specialized for output heads and does not compress transformer MLP or attention weights. Its benefit is largest when the vocabulary projection is both dense and sensitive to ordinary low-bit quantization; on Gemma-4-E4B, naive INT4 is already slightly better in the displayed rounded relative-PPL result. The cross-model study covers three quality evaluations, while packed size is checked on five heads; broader architectures, languages, context lengths, and calibration domains remain to be tested.

Our principal memory claim concerns serialized and load-time parameter tensors. Total peak GPU memory is workload- and backend-dependent because logits, activations, workspaces, allocator behavior, and KV caches can dominate. The three downstream tasks are only a sanity check and do not establish preservation of every model capability or safety property. Finally, the conditional optimality result applies to the unquantized rank-$r$ correction for a fixed core; quantizing its factors introduces additional approximation error, and the end-to-end ARCHead construction is not claimed to be globally optimal.

\section*{Ethical Considerations}

This work uses publicly released model checkpoints and WikiText-103; it does not introduce human-subject data collection or annotation. Compression can lower deployment costs and thereby broaden access, but it can also make models with existing biases or unsafe behaviors easier to deploy. ARCHead does not remove such behavior: a compressed model inherits the risks, licenses, and intended-use constraints of its source checkpoint. Practitioners should therefore repeat application-specific quality and safety evaluations after compression rather than relying only on perplexity or logit fidelity.

\paragraph{Acknowledgments.}
We thank TextCortex AI for providing GPU support for this study.

\bibliography{references}

\appendix
\input{appendix}

\end{document}

%% file: appendix.tex
\section{Extended Notation and Objective}
\label{app:extended_objective}

This section collects the notation used throughout the construction and makes explicit the relationship between head-weight error and logit error. Let $V$ denote vocabulary size, $D$ hidden width, and $N$ the number of calibration tokens. The dense LM-head is $\mathbf{W}\in\mathbb{R}^{V\times D}$, the calibration activations are $\mathbf{H}\in\mathbb{R}^{N\times D}$, and the compressed head is $\widehat{\mathbf{W}}$. Table~\ref{tab:app_notation} summarizes the remaining symbols.

\begin{table}[h]
\centering
\begin{tabular}{ll}
\toprule
Symbol & Meaning \\
\midrule
$\mathbf{W}_d$ & fixed packed quantized core \\
$\mathbf{E}$ & core residual $\mathbf{W}-\mathbf{W}_d$ \\
$\mathbf{C}_\lambda$ & damped activation covariance \\
$\mathbf{T}_p$ & activation-metric transform \\
$r_c,r_r$ & core and correction ranks \\
$g$ & quantization group size \\
$\mathbf{A}_w\mathbf{B}_w$ & low-rank residual correction \\
\bottomrule
\end{tabular}
\caption{Notation used in the ARCHead construction.}
\label{tab:app_notation}
\end{table}

For an arbitrary weight error $\boldsymbol{\Delta}=\mathbf{W}-\widehat{\mathbf{W}}$, the average squared logit error over the calibration activations is
\begin{align}
\frac{1}{N}\lVert\mathbf{H}\boldsymbol{\Delta}^{\top}\rVert_F^2
&=\frac{1}{N}\operatorname{Tr}\!\left(
\mathbf{H}\boldsymbol{\Delta}^{\top}\boldsymbol{\Delta}\mathbf{H}^{\top}
\right) \\
&=\operatorname{Tr}\!\left(
\boldsymbol{\Delta}\frac{\mathbf{H}^{\top}\mathbf{H}}{N}
\boldsymbol{\Delta}^{\top}\right).
\end{align}
Thus, directions with greater activation energy receive greater weight. Ordinary Frobenius reconstruction treats the covariance as the identity and cannot distinguish frequently activated directions from nearly inactive ones.

ARCHead uses the damped covariance
\begin{equation}
\mathbf{C}_\lambda=\frac{\mathbf{H}^{\top}\mathbf{H}}{N}
+\lambda\bar c\mathbf{I}
=\mathbf{Q}\boldsymbol{\Lambda}\mathbf{Q}^{\top},
\end{equation}
where $\bar c$ is the mean diagonal covariance. The transform $\mathbf{T}_p=\mathbf{Q}\boldsymbol{\Lambda}^{p}\mathbf{Q}^{\top}$ interpolates between an unweighted residual approximation at $p=0$ and the damped empirical logit-MSE geometry at $p=\tfrac12$. Values above one half emphasize dominant activation directions more strongly; the Qwen configuration selects $p=0.75$ on calibration data. Damping keeps the transform numerically stable in weakly observed directions and makes its inverse well defined.

\paragraph{Scope of the optimality statement.}
For a fixed core $\mathbf{W}_d$, invertible $\mathbf{T}_p$, and unquantized rank-$r_r$ correction, the truncated SVD of $\mathbf{E}\mathbf{T}_p$ is optimal in the induced metric. This statement does not optimize the core, metric power, or damping jointly. It also precedes INT8 factor quantization and therefore characterizes the ideal correction stage rather than the complete packed module. These qualifications are important when interpreting the theorem as an explanation of the correction design rather than as a claim of global end-to-end optimality.

\section{Expanded Construction Procedure}
\label{app:expanded_algorithm}

The construction consumes a dense head only while fitting the packed replacement. It proceeds as follows.

\begin{enumerate}
    \item \textbf{Collect head inputs.} Run calibration text through the model and retain final hidden states immediately before the LM-head. The evaluation activations used for reported quality metrics remain disjoint from calibration.
    \item \textbf{Estimate activation geometry.} Accumulate $\mathbf{H}^{\top}\mathbf{H}/N$, apply damping, and eigendecompose the resulting $D\times D$ matrix.
    \item \textbf{Construct the low-rank core.} Compute the rank-$r_c$ core factors and quantize the left and right factors with their designated precisions.
    \item \textbf{Quantize the core residual.} Subtract the unquantized low-rank core from $\mathbf{W}$ and encode the remaining matrix with signed group-wise INT4 values and quantized scales.
    \item \textbf{Measure the realized core error.} Dequantize the packed core representation for construction only and form $\mathbf{E}=\mathbf{W}-\mathbf{W}_d$. Using the realized core error ensures that the correction targets errors introduced by both approximation and factor quantization.
    \item \textbf{Fit in the activation metric.} Form $\mathbf{E}\mathbf{T}_p$, compute its randomized rank-$r_r$ truncated SVD, and map the right factor back through $\mathbf{T}_p^{-1}$.
    \item \textbf{Quantize and pack the correction.} Store $\mathbf{A}_w$ row-wise and $\mathbf{B}_w$ group-wise in INT8 together with their scales and shape metadata.
    \item \textbf{Discard the dense source.} Register only the packed tensors in \texttt{ARCHead}; the original BF16 $V\times D$ tensor and construction temporaries are not part of the serialized module.
\end{enumerate}

\begin{algorithm}[h]
\caption{Expanded packed-head construction}
\label{alg:app_expanded}
\begin{algorithmic}[1]
\REQUIRE $\mathbf{W}$, $\mathbf{H}$, $r_c$, $r_r$, group size $g$, $p$, $\lambda$
\ENSURE Packed module with no dense BF16 head
\STATE $\mathbf{C}_\lambda\leftarrow\mathbf{H}^{\top}\mathbf{H}/N+\lambda\bar c\mathbf{I}$
\STATE $(\mathbf{Q},\boldsymbol{\Lambda})\leftarrow\operatorname{Eigh}(\mathbf{C}_\lambda)$
\STATE $\mathbf{T}_p\leftarrow\mathbf{Q}\boldsymbol{\Lambda}^{p}\mathbf{Q}^{\top}$
\STATE $\mathbf{T}_p^{-1}\leftarrow\mathbf{Q}\boldsymbol{\Lambda}^{-p}\mathbf{Q}^{\top}$
\STATE $\mathbf{W}_d\leftarrow\operatorname{QuantizeCore}(\mathbf{W},r_c,g)$
\STATE $\mathbf{E}\leftarrow\mathbf{W}-\mathbf{W}_d$
\STATE $(\mathbf{U},\boldsymbol{\Sigma},\mathbf{V})\leftarrow
\operatorname{RandomizedSVD}(\mathbf{E}\mathbf{T}_p,r_r)$
\STATE $\mathbf{A}_w\leftarrow\mathbf{U}\boldsymbol{\Sigma}$
\STATE $\mathbf{B}_w\leftarrow\mathbf{V}^{\top}\mathbf{T}_p^{-1}$
\STATE $(\mathbf{A}_w^q,\mathbf{B}_w^q)\leftarrow
\operatorname{QuantizeINT8}(\mathbf{A}_w,\mathbf{B}_w)$
\RETURN $\operatorname{Pack}(\mathbf{W}_d,\mathbf{A}_w^q,\mathbf{B}_w^q,g)$
\end{algorithmic}
\end{algorithm}

At inference, ARCHead evaluates the core path and correction path separately,
\begin{equation}
\widehat{\mathbf{Y}}=mathbf{H}\mathbf{W}_d^{\top}
+(\mathbf{H}\mathbf{B}_w^{\top})\mathbf{A}_w^{\top}.
\end{equation}
This factorized evaluation avoids registering a reconstructed dense head. A backend may dequantize tiles or factors internally, but such temporary execution state is distinct from the persistent module representation measured in the storage tables.

\section{Packed Buffers and Model-Conversion Lifecycle}
\label{app:packed_lifecycle}

Table~\ref{tab:app_buffers} lists the logical components of the serialized module. Exact tensor layouts can vary with the packing implementation, but every reported byte count is obtained from the actual registered parameters and buffers rather than inferred solely from nominal bit widths.

\begin{table}[h]
\centering
\resizebox{\columnwidth}{!}{
\begin{tabular}{lll}
\toprule
Component & Stored form & Role \\
\midrule
Core left factor & packed 5-bit + scales & vocabulary-side core \\
Core right factor & packed INT8 + scales & hidden-side core \\
Core residual & signed packed INT4 & full-shape residual \\
Residual scales & quantized group scales & INT4 reconstruction \\
Correction left & row-wise INT8 + scales & vocabulary-side repair \\
Correction right & group-wise INT8 + scales & hidden-side repair \\
Metadata & integer shapes/groups & unpacking and dispatch \\
\bottomrule
\end{tabular}}
\caption{Logical contents of the packed \texttt{ARCHead} state dictionary.}
\label{tab:app_buffers}
\end{table}

\paragraph{Conversion lifecycle.}
The source checkpoint is first loaded with its dense LM-head available to the constructor. Calibration activations and the metric are computed, the quantized core and correction are fitted, and a packed \texttt{ARCHead} instance is created. The model's output embedding is then replaced by this module. Before serialization, the dense source tensor and temporary decompositions are released. Reloading the converted checkpoint instantiates only the packed buffers listed above.

\paragraph{Persistent storage versus working memory.}
Persistent storage is the sum of \texttt{numel}$\times$\texttt{element\_size} over registered tensors. Construction working memory can be larger because it includes $\mathbf{W}$, covariance factors, randomized-SVD workspaces, and temporary dequantized values. Forward working memory is different again and can include logits, activations, the KV cache, allocator state, and backend workspaces. The 3.7--3.9$\times$ claim concerns the first quantity only; separating these categories prevents a packed-parameter result from being misread as an identical reduction in end-to-end peak GPU memory.

\paragraph{Integration order.}
ARCHead is applied after the chosen transformer-block quantizer. This order preserves the block backend and targets only the remaining dense output projection. In the inspected Qwen AWQ and bitsandbytes checkpoints, the head remained a BF16 \texttt{Linear} of shape $151{,}936\times4{,}096$, so replacement does not require changing the transformer-block representation.

\section{Evaluation and Reproducibility Checklist}
\label{app:evaluation_checklist}

The following checklist clarifies how the reported comparisons isolate the output head.

\begin{itemize}
    \item \textbf{Head-only experiments:} transformer blocks remain BF16 and only the LM-head representation changes.
    \item \textbf{Hybrid experiments:} AWQ or bitsandbytes quantizes the blocks first; BF16-head and ARCHead rows share the same block backend.
    \item \textbf{Calibration split:} final hidden states are sampled from the WikiText-103 training split.
    \item \textbf{Evaluation split:} quality is measured on 16,384 held-out test tokens that are not reused for fitting.
    \item \textbf{Matched seeds:} the GPTQ-style head and ARCHead use the same activation cache at each 2K, 4K, 8K, and 16K calibration budget with seeds 0, 1, and 2.
    \item \textbf{Storage measurement:} bytes are summed from instantiated packed tensors, including scales and metadata-bearing buffers.
    \item \textbf{Timing comparison:} paired construction times use the same NVIDIA RTX Pro 6000; speedup is GPTQ time divided by ARCHead time.
    \item \textbf{Throughput interpretation:} changes within 2\% are treated as negligible measurement variation, and small positive values are not claimed as speedups.
\end{itemize}

\paragraph{Metric interpretation.}
Cross-entropy and relative perplexity measure language-model quality under the fixed evaluation activations. Logit MSE and KL divergence measure fidelity to the dense head, while top-$k$ agreement records whether dense predictions remain among the compressed head's leading candidates. Persistent head ratio measures serialized module size relative to BF16. These metrics answer different questions and should not be collapsed into a single claim: storage does not determine quality, logit fidelity does not guarantee every downstream capability, and persistent bytes do not equal runtime peak memory.

\paragraph{Downstream sanity check.}
HellaSwag, TruthfulQA MC2, and WinoGrande are evaluated with fixed transformer blocks and evaluation settings across head variants. The suite is intentionally described as a sanity check because its small differences do not establish task-level improvement or comprehensive capability preservation. Its role is to identify an obvious regression that might be hidden by aggregate perplexity alone.

\section{Full Head-Only Results}

Table \ref{tab:app_head_only} provides the comprehensive head-only quantization results for Qwen3-8B-Base. In these experiments, the transformer blocks are deliberately kept in BF16 to isolate the performance impact of the LM-head. As the results indicate, ARCHead achieves a storage footprint similar to naive Group INT4, but with significantly lower cross-entropy and perplexity degradation.

\begin{table}[h]
\centering
\resizebox{\columnwidth}{!}{
\begin{tabular}{lrrrrr}
\toprule
\textbf{Method} & \textbf{Head Ratio} & \textbf{CE} & \textbf{Delta CE} & \textbf{PPL} & \textbf{Relative PPL} \\
\midrule
Dense BF16 & 1.0000 & 2.443 & 0.000 & 11.504 & 1.000 \\
Row INT8 & 0.5000 & 2.460 & +0.017 & 11.704 & 1.017 \\
Group INT4 & 0.2656 & 2.583 & +0.140 & 13.237 & 1.151 \\
SVD8 + INT4 & $\sim$0.2700 & 2.634 & +0.191 & 13.929 & 1.211 \\
\textbf{ARCHead} & 0.2557 & 2.450 & +0.007 & 11.588 & 1.007 \\
\bottomrule
\end{tabular}
}
\caption{Full head-only quantization results on Qwen3-8B-Base. Transformer blocks remain in BF16.}
\label{tab:app_head_only}
\end{table}

\section{Cross-Model Head-Only Results}

Table \ref{tab:app_cross_model} shows the performance of ARCHead across models with differing vocabularies and hidden dimensions. 

Qwen3-8B exhibits the strongest relative improvement because its naive INT4 baseline fails drastically. For Gemma-4-E4B, the naive INT4 baseline is already strong, meaning ARCHead's advantage is narrower, though it remains highly competitive. VibeThinker-3B shows a larger overall loss due to its smaller hidden dimension, but ARCHead still provides a substantial improvement over the naive INT4 baseline. This highlights the model-dependent behavior of LM-head quantization sensitivity.

\begin{table}[h]
\centering
\resizebox{\columnwidth}{!}{
\begin{tabular}{lrrrrr}
\toprule
\textbf{Model} & \textbf{Dense PPL} & \textbf{INT4 Rel PPL} & \textbf{ARCHead Rel PPL} & \textbf{ARCHead Delta CE} & \textbf{Head Ratio} \\
\midrule
Qwen3-8B & 11.504 & 1.151 & 1.007 & +0.007 & 0.256 \\
Gemma-4-E4B & 11.970 & 1.013 & 1.010 & +0.010 & 0.269 \\
VibeThinker-3B & 12.450 & 1.050 & 1.027 & +0.026 & 0.257 \\
\bottomrule
\end{tabular}
}
\caption{Cross-model head-only results demonstrating model-dependent sensitivity.}
\label{tab:app_cross_model}
\end{table}

\section{Hybrid Quantizer + ARCHead Results}

ARCHead is intended as a drop-in replacement for backends like AWQ and bitsandbytes. Table \ref{tab:app_hybrid} presents the hybrid setup. ARCHead does not improve the base perplexity of AWQ or BNB; rather, it compresses the dense BF16 head they leave behind. The additional cross-entropy cost introduced by ARCHead is extremely small ($+0.006$ to $+0.007$), validating the drop-in replacement claim.

\begin{table}[h]
\centering
\resizebox{\columnwidth}{!}{
\begin{tabular}{llrrrr}
\toprule
\textbf{Block Quantizer} & \textbf{LM-Head} & \textbf{Head Ratio} & \textbf{Rel PPL} & \textbf{Delta CE} & \textbf{Extra CE} \\
\midrule
AWQ 4-bit & Dense BF16 & 1.000 & 1.0465 & +0.0454 & 0.000 \\
AWQ 4-bit & \textbf{ARCHead} & 0.256 & 1.0530 & +0.0516 & +0.006 \\
BNB NF4 & Dense BF16 & 1.000 & 1.1084 & +0.1030 & 0.000 \\
BNB NF4 & \textbf{ARCHead} & 0.256 & 1.1170 & +0.1100 & +0.007 \\
\bottomrule
\end{tabular}
}
\caption{Hybrid model results on Qwen3-8B-Base. ARCHead compresses the remaining dense head with minimal extra CE.}
\label{tab:app_hybrid}
\end{table}

\section{Ablation Details}

The extended ablation study in Table \ref{tab:app_ablation} underscores the necessity of the activation-metric residual correction branch. The ``ARCHead core only'' variant exhibits a relative PPL of 1.134. Adding the residual correction branch reduces this sharply to 1.007. This confirms that the core alone is insufficient and supports the central novelty of optimizing in the covariance-weighted logit space.

\begin{table}[h]
\centering
\resizebox{\columnwidth}{!}{
\begin{tabular}{lrrrrr}
\toprule
\textbf{Variant} & \textbf{Head Ratio} & \textbf{CE} & \textbf{Delta CE} & \textbf{PPL} & \textbf{Rel PPL} \\
\midrule
Dense BF16 & 1.000 & 2.443 & 0.000 & 11.504 & 1.000 \\
Group INT4 & 0.266 & 2.583 & +0.140 & 13.237 & 1.151 \\
SVD8 + INT4 & $\sim$0.270 & 2.634 & +0.191 & 13.929 & 1.211 \\
ARCHead core only & $\sim$0.230 & 2.569 & +0.126 & 13.048 & 1.134 \\
\textbf{ARCHead full} & 0.256 & 2.450 & +0.007 & 11.588 & 1.007 \\
\bottomrule
\end{tabular}
}
\caption{Expanded ablation details highlighting the importance of the correction branch.}
\label{tab:app_ablation}
\end{table}

\section{Throughput Measurement Details}

Table \ref{tab:app_throughput} outlines generation throughput measurements. ARCHead does not bottleneck generation. The small positive difference observed in the BNB NF4 configuration represents measurement noise rather than a real speedup. Throughput is measured independently from persistent memory to provide an accurate picture of inference viability.

\begin{table}[!ht]
\centering
\resizebox{\columnwidth}{!}{
\begin{tabular}{lrrr}
\toprule
\textbf{Model / Backend} & \textbf{BF16 Head tok/s} & \textbf{ARCHead tok/s} & \textbf{Change} \\
\midrule
Qwen3-8B Dense & 4468 & 4467 & negligible \\
Qwen3-8B AWQ & 3452 & 3451 & negligible \\
Qwen3-8B BNB NF4 & 1789 & 1813 & meas. noise / no slowdown \\
VibeThinker-3B Dense & 7091 & 6947 & $\sim-$2\% \\
\bottomrule
\end{tabular}
}
\caption{ARCHead maintains generation throughput.}
\label{tab:app_throughput}
\end{table}

\FloatBarrier
\section{Additional Result Visualizations}

\begin{figure}[!ht]
\centering
\includegraphics[width=1.0\columnwidth]{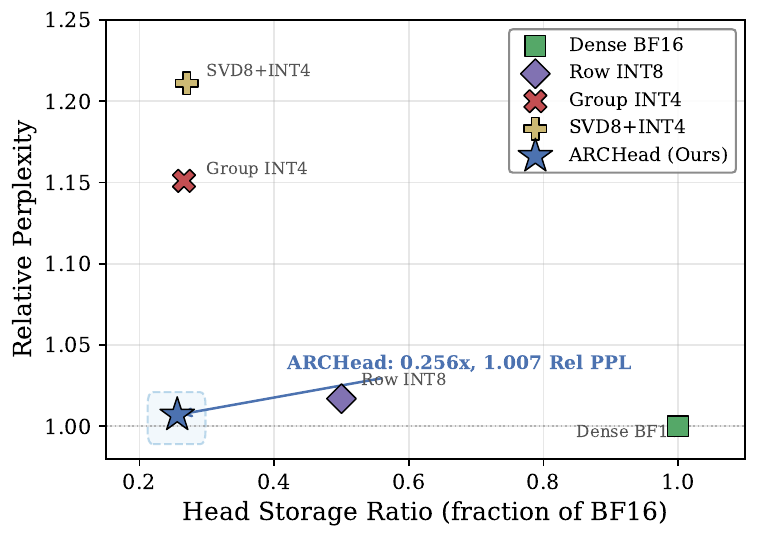}
\caption{Qwen3-8B-Base head-storage and perplexity trade-off. ARCHead
occupies a similar storage range to group INT4 while remaining close to
the dense head.}
\label{fig:app_tradeoff}
\end{figure}

\begin{figure}[!ht]
\centering
\includegraphics[width=1.0\columnwidth]{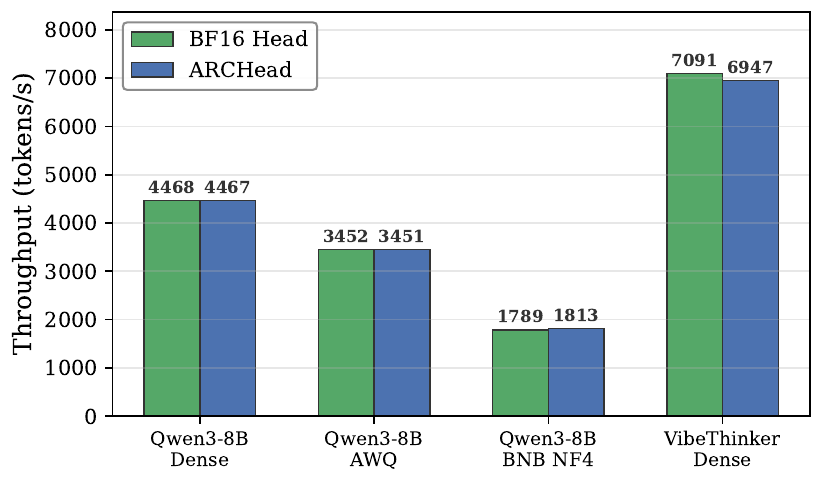}
\caption{Measured generation throughput with a BF16 or packed ARCHead
head. Differences below 2\% are treated as negligible measurement
variation.}
\label{fig:app_throughput}
\end{figure}

\begin{figure}[!ht]
\centering
\includegraphics[width=1.0\columnwidth]{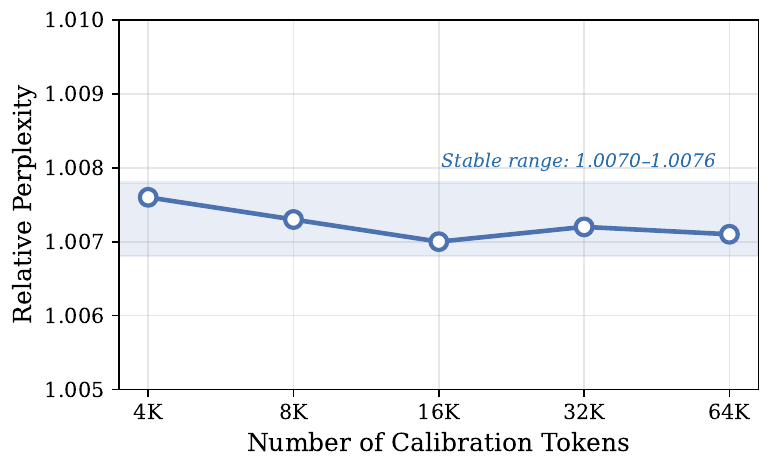}
\caption{Qwen3-8B-Base calibration sensitivity. Relative PPL remains in
a narrow band from 4K through 64K calibration tokens.}
\label{fig:app_calibration}
\end{figure}

\FloatBarrier
\section{Failure Modes and Limitations}

The scope and limitations of ARCHead include:
\begin{itemize}
    \item \textbf{Density Requirement:} ARCHead is most beneficial when the LM-head constitutes a large dense parameter block.
    \item \textbf{Model Sensitivity:} If a model's naive INT4 head quantization already performs exceptionally well (as observed to some degree with Gemma), ARCHead's relative advantage diminishes.
    \item \textbf{Prior Compression:} If a backend natively and successfully quantizes the LM-head, applying ARCHead may not yield additional quality improvements.
    \item \textbf{Scope:} ARCHead is explicitly developed for LM-head compression. It is not intended for compressing Feed-Forward Networks (FFN/MLP) in this paper.
    \item \textbf{Evaluation:} WikiText perplexity and logit fidelity are the primary quality measures. The three downstream tasks are a limited sanity check; exhaustive capability and safety evaluation remains future work.
    \item \textbf{Kernel Optimization:} The included fused-kernel result is preliminary and limited to one accelerator. Broader hardware and production-backend optimization remains future work.
\end{itemize}

\section{Preliminary Future Work: Fused Triton Kernel Integration}
\label{sec:app_future_kernel}

We present preliminary results from a custom Triton kernel designed for
the ARCHead compressed head. The kernel performs INT8 dequantization and
evaluates the low-rank residual branch within a fused pass, loading the
quantized core and row-wise scales directly while computing the
low-rank residual.

This prevents the prohibitive memory bandwidth overhead of sequentially instantiating the full FP16 matrix before multiplication. Table \ref{tab:app_triton_throughput} presents the throughput results of integrating this fused kernel on a single NVIDIA RTX Pro 6000. 

These preliminary findings confirm that ARCHead does not bottleneck inference generation in practical deployment scenarios, even when paired with aggressively quantized transformer backends like AWQ or bitsandbytes NF4.

\newpage
\begin{table}[!h]
\centering
\resizebox{\columnwidth}{!}{
\begin{tabular}{lrrr}
\toprule
\textbf{Model / Backend} & \textbf{BF16 Head (tok/s)} & \textbf{ARCHead Fused Kernel (tok/s)} & \textbf{Change} \\
\midrule
Qwen3-8B Dense & 4404.8 & 4398.0 & $\sim-$0.15\% \\
Qwen3-8B BNB NF4 & 1803.7 & 1814.6 & $\sim+$0.60\% \\
Qwen3-8B AWQ 4-bit & 3468.4 & 3490.5 & $\sim+$0.60\% \\
VibeThinker-3B Dense & 7011.9 & 6883.1 & $\sim-$1.80\% \\
VibeThinker-3B BNB NF4 & 3192.6 & 3198.0 & $\sim+$0.10\% \\
VibeThinker-3B AWQ 4-bit & 4510.8 & 4508.4 & $\sim-$0.05\% \\
\bottomrule
\end{tabular}
}
\caption{Preliminary throughput validation of the ARCHead fused Triton
kernel on one NVIDIA RTX Pro 6000. The measured differences are within
2\%; small positive values are treated as noise rather than speedups.}
\label{tab:app_triton_throughput}
\end{table}